\documentclass[lnbip]{svmultln}
\usepackage{makeidx}  

\usepackage{amssymb}
\usepackage{adjustbox}
\usepackage{float}
\usepackage{color,soul}
\usepackage{listings}
\usepackage{xcolor}

\usepackage{tikz,xcolor,hyperref}

\definecolor{lime}{HTML}{A6CE39}
\DeclareRobustCommand{\orcidicon}{%
    \begin{tikzpicture}
    \draw[lime, fill=lime] (0,0) 
    circle [radius=0.16] 
    node[white] {{\fontfamily{qag}\selectfont \tiny ID}};
    \draw[white, fill=white] (-0.0625,0.095) 
    circle [radius=0.007];
    \end{tikzpicture}
    \hspace{-2mm}
}

\newcommand{\orcidJMR}{\href{https://orcid.org/0000-0002-3665-639X}{\orcidicon}}
\newcommand{\orcidPZ}{\href{https://orcid.org/0000-0002-6630-3106}{\orcidicon}}
\newcommand{\orcidKK}{\href{https://orcid.org/0000-0002-4918-0650}{\orcidicon}}
\newcommand{\orcidIN}{\href{https://orcid.org/0000-0003-2598-0116}{\orcidicon}}
\newcommand{\orcidBF}{\href{https://orcid.org/0000-0002-8585-9388}{\orcidicon}}
\newcommand{\orcidDM}{\href{https://orcid.org/0000-0003-4480-082X}{\orcidicon}}

\restylefloat{table}

\lstdefinestyle{shared}{
    breaklines=true,
    tabsize=2,
    columns=flexible,
    basicstyle=\fontsize{10}\selectfont,
    backgroundcolor = \color{codebkg},
    breakatwhitespace=false
}
\begin{document}
\mainmatter 
\title{XAI-KG: knowledge graph to support XAI and decision-making in manufacturing.}
\titlerunning{Knowledge graph to support XAI and decision-making in manufacturing.}  
%
\author{Jo\v{z}e M. Ro\v{z}anec*\inst{1,2,3} \orcidJMR \and Patrik Zajec\inst{1,2} \orcidPZ \and 
Klemen Kenda\inst{1,2,3} \orcidKK \and Inna Novalija\inst{2} \orcidIN \and Bla\v{z} Fortuna\inst{2,3} \orcidBF \and Dunja Mladeni\'{c}\inst{2} \orcidDM}

\authorrunning{Jo\v{z}e M. Ro\v{z}anec et al.}   
%
\tocauthor{Jo\v{z}e M. Ro\v{z}anec, Patrik Zajec, Klemen Kenda, Inna Novalija, Bla\v{z} Fortuna, Dunja Mladeni\'{c}}
\institute{Jo\v{z}ef Stefan International Postgraduate School, Jamova 39, 1000 Ljubljana, Slovenia,\\
\email{joze.rozanec@ijs.si},
\and
Jo\v{z}ef Stefan Institute, Jamova 39, 1000 Ljubljana, Slovenia
\and 
Qlector d.o.o., Rov\v{s}nikova 7, 1000 Ljubljana, Slovenia}

\maketitle              

\begin{abstract}        
The increasing adoption of artificial intelligence requires accurate forecasts and means to understand the reasoning of artificial intelligence models behind such a forecast. Explainable Artificial Intelligence (XAI) aims to provide cues for why a model issued a certain prediction. Such cues are of utmost importance to decision-making since they provide insights on the features that influenced most certain forecasts and let the user decide if the forecast can be trusted. Though many techniques were developed to explain black-box models, little research was done on assessing the quality of those explanations and their influence on decision-making. We propose an ontology and knowledge graph to support collecting feedback regarding forecasts, forecast explanations, recommended decision-making options, and user actions. This way, we provide means to improve forecasting models, explanations, and recommendations of decision-making options. We tailor the knowledge graph for the domain of demand forecasting and validate it on real-world data.
\end{abstract}
%


\section{Introduction}\label{INTRODUCTION}
The increasing digitization of manufacturing enables data-driven decision-making and wider adoption of Artificial Intelligence (AI). AI models can learn from past data and predict future outcomes. Though making forecasts more accurate is important, the lack of transparency on how such forecasts are achieved can undermine trust in the AI models. To mitigate this issue, techniques were developed to identify which variables were most influential to a given forecast\cite{adadi2018peeking,li2020survey}. While such information is valuable, it is equally important to convey such information appealingly and clearly to the user\cite{pedreschi2018open}. Though much research focused on techniques to identify influential forecasts, there are few validated measurements for user evaluations on explanations' quality and how do such explanations influence decision-making\cite{van2021evaluating,zhou2021evaluating}.

Of central concern to achieve user evaluations on explanations' quality and their influence on decision-making is collecting, modeling, and storing related data. Such data are the predictive models, their forecasts, the main features influencing such forecasts, the use cases they relate to, and users' feedback. Such data can be used to complete the missing information in the knowledge graph and assess the quality of current forecast explanations. It also provides means to enhance future models' explanations and decision-making options presented to the user.

Though AI models and decision-making options differ between the different use cases, there are many common entities. We consider a generic knowledge graph to support collecting user feedback on given model forecast explanations, how those explanations influenced decision-making, and users' decisions while encoding meaningful context.

This research develops an ontology and knowledge graph for manufacturing to collect feedback regarding forecasts, XAI, and decision-making options. We also design a domain-specific ontology, which provides relevant concepts that must be taken into account to create such a knowledge graph regardless of the use case. Finally, we test our approach for the domain of demand forecasting and validate it on a real-world case study, using models we developed as part of the European Horizon 2020 projects FACTLOG\footnote{https://www.factlog.eu/} and STAR\footnote{http://www.star-ai.eu/}.

As part of this research, we published an ontology with the concepts and relationships we considered to build the knowledge graph. To provide insights regarding the dimension and interconnectedness of the knowledge graph, we computed a set of metrics presented in Section~\ref{RESULTS-AND-EVALUATION}.

The rest of this paper is structured as follows: Section~\ref{RELATED-WORK} presents related work, Section~\ref{ONTOLOGY-AND-KB} describes the ontology we used and evolved, and the knowledge graph we designed, Section~\ref{USE-CASE} describes the use case we used and implementation we created to test our concept, Section~\ref{RESULTS-AND-EVALUATION} provides the results we obtained and evaluates the knowledge graph. Finally, in Section~\ref{CONCLUSION}, we provide our conclusions and outline future work.

\section{Related Work}\label{RELATED-WORK}
The increasing digitalization of the manufacturing domain\cite{xu2018industry} enables the development and usage of knowledge graphs in manufacturing. \cite{buchgeher2020knowledge} found that knowledge graphs are most required for knowledge fusion and frequently considered enablers for other use cases. Ontologies can be used to guide the knowledge graph construction, helping to define formal terminology and transfer knowledge into graphical form\cite{8777086}. Though many sources of data exist, conversational interfaces provide a flexible medium to gather locally observed collective knowledge\cite{preece2015sherlock,bradevsko2017curious}. Though approaches were developed to query existing knowledge in manufacturing systems\cite{bunte2016integrating}, we found no literature using such interfaces to gather data in a manufacturing setting. Such an interface can be useful to assess the quality of forecast explanations\cite{10.1145/3432934}, extended to decision-making options, and help enhance their quality over time.

In this research, we dedicate particular attention to integrating knowledge modeling techniques in manufacturing with XAI, decision-making options, persisting knowledge, and feedback we aim to obtain through a question-answering interface.

\section{Ontology and Knowledge Graph}\label{ONTOLOGY-AND-KB}
In our research, the ontology defines important concepts that later guide the development of a knowledge graph. When developing the ontology, we defined its scope and level of formality required\cite{uschold1996building,kim1995ontology}. In particular, we require a domain-specific ontology that supports conversational interfaces to collect knowledge regarding manufacturing processes and feedback regarding explanations provided for AI models forecasts (XAI) and decision-making options. Such a knowledge graph provides ground to assess users' perceived quality of forecast explanations, a yet little researched aspect in XAI\cite{van2021evaluating}. To reuse existing ontology concepts\cite{uschold1995towards,fernandez1997methontology}, we use Basic Formal Ontology (BFO)\cite{Smith2002} and Industrial Ontologies Foundry (IOF) specifications\cite{Kulvatunyou2018} and import concepts from \cite{DVN/DVZH81_2021}. Among new concepts we introduced in our ontology are \textit{Feature relevance}, \textit{Forecast explanation}, and \textit{Feedback}. We implemented the Knowledge Graph using Neo4j\footnote{https://neo4j.com/}. We developed an ontology with general concepts that can be reused in any use case, and an extended version with concepts related to the demand forecasting use case\footnote{We published both ontologies at the Harvard Dataverse. They are accessible from the following link: \url{https://doi.org/10.7910/DVN/UGYHLP}}.

\section{Use case}\label{USE-CASE}
To ensure our knowledge graph meets the goals described in Section~\ref{INTRODUCTION}, we instantiated it for the demand forecasting use case with data obtained from partners related to the aforementioned EU Horizon 2020 projects. Demand forecasting is a critical component of supply change management. It guides operational and strategic decisions regarding resources, workers, manufactured products, and logistics. The increasing digitalization and information sharing in the manufacturing sector accelerate the information flow.  It provides means to create frequent and timely forecasts, which can consider the latest context, improving the quality of forecasts. Though increasing the quality of forecasts will always remain a priority, it is equally important to provide some forecast explanations. Such explanations provide insights into elements relevant to specific forecasts and allow the user to judge if such a forecast can be trusted. Forecasts and forecast explanations are needed to guide decision-making. Forecasts provide ground to suggest decision-making options to the users, which can alleviate user decision-making over time. Collecting feedback on provided forecasts, explanations, and decision-making options is of utmost importance to improve forecasting models, XAI techniques, forecast explanations display, and recommended decision-making options.

For this research, we created an ontology and implemented a knowledge graph following concepts and relationships elicitated in the ontology. The knowledge graph was implemented with the Neo4j graph database. Ingested data included three years of shipment information daily, a month of demand forecasts for material and clients at a daily level, feature relevance for every forecast (computed with LIME\cite{ribeiro2016should} library), forecast explanations created based on those feature rankings, and decision-making options created based on demand forecasts and heuristics. We did not collect feedback from the users. However, we created synthetic data to simulate expected feedbacks on forecasts, decision-making options, and feature relevance obtained from the LIME library.

\section{Results}\label{RESULTS-AND-EVALUATION}
We developed and published a domain-specific ontology that aims to represent relevant entities regarding forecasts, their explanations, decision-making options, and feedback we collect. We also published an extended version, in which we include concepts related to the demand forecasting use case.

To assess the knowledge graph structure, we adopted five metrics suggested in \cite{mathieson2010complexity}: \# nodes (number of nodes), \# paths (number of paths), Total Path Lenght (TPL - the sum of relationships traversed while traveling from each node to every other node), Maximum Path Length (MPL - the maximum length among shortest paths between the nodes), and Average Path Length (APL - the average of shortest path lengths between nodes). The number of nodes and relationships provide an insight into the Knowledge Graph dimensions, while TPL, MPL, and APL measure the nodes' interconnectedness. We measured 80.948 nodes and 156.485 paths in our implementation. To estimate the rest of the metrics, we randomly sampled 0,05\% of graph nodes and paths, and measured TPL: 435.943.206, MPL: 8, and APL: 5,99.

\section{Conclusions}\label{CONCLUSION}
The increasing level of digitalization and AI adoption in manufacturing require developing mechanisms to provide users insights into reasons driving AI model forecasts. An increasing body of research is developing mechanisms to provide transparency to black-box models. In addition to these techniques, users' feedback is of utmost importance to understand how explanations influence users' behavior and improvement opportunities. Though much information regarding the manufacturing process is captured by software such as Enterprise Resource Planning or Manufacturing Execution System platforms, specific details cannot be captured through regular interfaces. The use of conversational interfaces can help bridge the gap, collecting situational knowledge that is otherwise lost. Such knowledge can provide insights into demand forecasting model biases and guide further development of XAI by providing insight into the perceived value of forecast explanations and their influence on decision-making, guide further development of decision-making recommendations and ranking, and help enrich the knowledge graph with locally observed collective knowledge. In particular, for demand forecasting, such enrichment involves registering new decision-making options, provide knowledge regarding logistics (transport types, pallet sizes, rationales behind logistics decision-making, and other data and criteria that are missing in the knowledge graph).

In this work, we introduced a domain-specific ontology and built a knowledge graph to support storing and relating feedback of explanations and recommended decision-making options and collect feedback of the users' actions. We tested our approach on the demand forecasting use case with real-world data. We demonstrate how the knowledge graph can support the aforementioned requirements, benefiting from the data's semantic representation.

As future work, we will use this knowledge graph to support and develop services with conversational interfaces to collect feedback regarding AI model forecasts and their explanations and decision-making options. We envision developing an active learning module, which will guide conversations to acquire meaningful and yet unlabeled data labels. Finally, data acquired through these services will develop AI-based recommender systems for decision-making options and research users' perception of XAI and its influence on decision-making.

\section*{Acknowledgement}
This work was supported by the Slovenian Research Agency and the European Union’s Horizon 2020 program projects FACTLOG under grant agreement H2020-869951, and STAR under grant agreement number H2020-956573.

%
%
\bibliographystyle{spmpsci}
\bibliography{main}
\end{document}